\documentclass{article}

\usepackage{amsfonts,amsmath,amssymb}
\usepackage{graphicx}
\usepackage{hyperref}
\usepackage{url}

\newtheorem{example}{Example}[section]
\newtheorem{claim}[example]{Claim}
\newtheorem{theorem}[example]{Theorem}
\newtheorem{proposition}[example]{Proposition}
\newtheorem{lemma}[example]{Lemma}
\newtheorem{corollary}[example]{Corollary}

\newenvironment{pf}{\emph{Proof:}}{}

\newenvironment{pfclaim}{\emph{Proof of Claim:}}{}

\newcommand{\qed}{\hfill Q.E.D.}

\newcommand{\fuzzyg}[2]{\mbox{$\langle #1 \geq #2\rangle$}}
\newcommand{\fuzzyneg}[2]{\mbox{$\langle #1 \leq #2\rangle$}}

\newcommand{\tuple}[1]{\langle #1 \rangle}

\newcommand{\T}{\mathcal{T}}

\newcommand{\A}{\mathcal{A}}
\newcommand{\KB}{\mathcal{K}}
\newcommand{\K}{\mathcal{K}_2}
\newcommand{\andc}{\sqcap}
\newcommand{\all}{\forall}
\newcommand{\some}{\exists}
\newcommand{\notc}{\neg}
\newcommand{\orc}{\sqcup}
\newcommand{\csome}{\exists}
\newcommand{\bottomc}{\perp}
\newcommand{\topc}{\top}
\newcommand{\impc}{\sqsubseteq}
\newcommand{\cass}[2]{\mbox{$#1\colon \!\!#2$}}
\newcommand{\rass}[3]{\mbox{$(#1,#2)\colon \!\!#3$}}
\newcommand{\AC}{{\mathbf C}}
\newcommand{\AR}{{\mathbf{R}}}
\newcommand{\AI}{{\mathbf{I}}}
\newcommand{\I}{\mathcal{I}}
\newcommand{\highi}[1]{{#1}^\I}

\newcommand{\unit}{[0,1]}
\newcommand{\ass}{{\mbox{$\tau$}}}
\newcommand{\notf}{\ominus}
\newcommand{\orf}{\oplus}
\newcommand{\andf}{\otimes}
\newcommand{\impf}{\Rightarrow}

\newcommand{\nd}{\noindent}

\begin{document}

\title{On the Failure of the Finite Model Property in some Fuzzy Description Logics}

\author{$^{1}$Fernando Bobillo \and  $^{2}$F\'{e}lix Bou \and $^{3}$Umberto Straccia \\ \\
$^{1}$ {\small Department of Computer Science and Systems Engineering} \\
{\small University of Zaragoza, Spain} \\ \\
$^{2}$ {\small Institut d'Investigaci\'{o} en Intellig\`{e}ncia Artificial}  \\
{\small Consejo Superior de Investigaciones Cient\'{\i}ficas (IIIA - CSIC), Bellaterra, Spain} \\ \\
$^{3}$ {\small Istituto di Scienza e Tecnologie dell'Informazione} \\
{\small Consiglio Nazionale delle Ricerche (ISTI - CNR), Pisa, Italy} 
}





\maketitle

\begin{abstract}
Fuzzy Description Logics (DLs) are a family of logics which allow
the representation of (and the reasoning with) structured knowledge
affected by vagueness. Although most of the not very expressive
crisp DLs, such as $\mathcal{ALC}$, enjoy the Finite Model Property
(FMP), this is not the case once we move into the fuzzy case. In
this paper we show that if we allow arbitrary knowledge bases, then
the fuzzy DLs $\mathcal{ALC}$ under \L ukasiewicz and Product fuzzy
logics do not verify the FMP even if we restrict to witnessed
models; in other words, finite satisfiability and witnessed
satisfiability are different for arbitrary knowledge bases. The aim
of this paper is to point out the failure of FMP because it affects
several algorithms published in the literature for reasoning under
fuzzy $\mathcal{ALC}$.
\end{abstract}




\section{Introduction}
~\label{sec:intro}

\emph{Description Logics} (DLs)~\cite{DLHandbook} are a logical
reconstruction of the so-called frame-based knowledge representation
languages, with the aim of providing a simple well-established
Tarski-style declarative semantics to capture the meaning of the
most popular features of structured representation of knowledge.
Nowadays, DLs have gained even more popularity due to their
application in the context of the Semantic
Web~\cite{DLasOLanguages}. For example, the current standard
language for specifying ontologies, the Web Ontology Language OWL is
based on Description Logics.

It is very natural to extend DLs to the fuzzy case in order to
manage fu\-zzy/va\-gue/im\-precise pieces of knowledge for which a
clear and precise definition is not possible. For a good and recent
survey on the advances in the field of fuzzy DLs, we refer the
reader to~\cite{StracciaSurvey}. One of the challenges of the
research in this community is the fact that different families of
fuzzy operators (or fuzzy logics) lead to fuzzy DLs with different
properties.

In fuzzy logic, there are a lot of families of fuzzy operators (or
fuzzy logics). Table~\ref{tab:fuzzyOperators} shows the connectives
involved in what are considered the main four families.
The most famous families correspond to the three basic continuous
t-norms (i.e., {\L}ukasiewicz, G\"odel and Product~\cite{Hajek98}) together with an
R-implication\footnote{Every
left-continuous t-norm $\otimes$ has associated a unique operation
$\Rightarrow$ called the residuum of $\otimes$ (often called an
R-implication) and defined as $\alpha \Rightarrow \beta = \sup\,\{
\gamma \mid \alpha \otimes \gamma \leq \beta \}$.}. Besides
these three, we also point out Zadeh's family corresponding to the
operators originally proposed by Lotfi A. Zadeh in his seminal
work~\cite{Zadeh65}: G{\"o}del t-norm and t-conorm, \L{}ukasiewicz
negation and Kleene-Dienes implication.

\begin{table}[htbp]
\caption{Popular families of fuzzy operators}
\label{tab:fuzzyOperators}
\begin{scriptsize}
\begin{center}
\begin{tabular}{|l|l|l|l|l|}
    \hline
    family & t-norm $\alpha \otimes \beta$ & t-conorm $\alpha \oplus \beta$ & negation $\ominus \alpha$ & implication $\alpha \Rightarrow \beta$ \\
    \hline \hline

    Zadeh & $\min \{\alpha, \beta\}$ & $\max \{\alpha, \beta\}$
    & $1 - \alpha$ & $\max \{1 - \alpha, \beta\}$ \\

    \hline

    {\L}ukasiewicz & $\max \{\alpha + \beta - 1, 0\}$ & $\min \{\alpha + \beta, 1\}$ & $1 - \alpha$ &
$\min \{1 - \alpha + \beta, 1\}$ \\

    \hline

    Product & $\alpha \cdot \beta$ & $\alpha + \beta - \alpha \cdot \beta$ &
    $\left\{
    \begin{array}{ll}
    1, & \alpha = 0 \\
    0, & \alpha > 0 \\
    \end{array}
    \right.$ &
    $\left\{
    \begin{array}{ll}
    1, & \alpha \leq \beta \\
    \beta / \alpha, & \alpha > \beta \\
    \end{array}
    \right.$\\

    \hline

    G\"{o}del & $\min \{\alpha, \beta\}$ & $\max \{\alpha, \beta\}$ &
    $\left\{
    \begin{array}{ll}
    1, & \alpha = 0 \\
    0, & \alpha > 0 \\
    \end{array}
    \right.$ &
    $\left\{
    \begin{array}{ll}
    1 & \alpha \leq \beta \\
    \beta, & \alpha > \beta \\
    \end{array}
    \right.$\\

    \hline
\end{tabular}
\end{center}
\end{scriptsize}
\end{table}

The DL $\mathcal{ALC}$~\cite{Schmidt91} ($\mathcal{A}$ttributive
$\mathcal{L}$anguage with $\mathcal{C}$omplements) is a very popular
logic, usually considered as a standard. $\mathcal{ALC}$ is a
notational variant of the multimodal logic $\mathbf{K_n}$ (which
essentially corresponds to not having a knowledge
base)~\cite{Schild91}, and it is well known that $\mathbf{K_n}$
verifies the Finite Model Property (FMP) (see~\cite[Proposition
2.15]{ModalLogics}). A logic verifies the FMP iff every satisfiable
theory of the logic is also satisfiable by a finite model.

However, the situation is different in the fuzzy setting. It is also
well known (see~\cite[Theorem~2]{Hajek05}
and~\cite[Theorem~8]{Hajek06b}\footnote{The result is formulated in
a different framework (for example, without explicitly having a
t-conorm), but the same proof given there works.}) that in the
context of fuzzy $\mathcal{ALC}$ the satisfiability problem for
ABoxes (that is, the TBox is empty) in a finite model coincides with
the satisfiability in the bigger class of witnessed models. This
fact is used in~\cite{Hajek05} to prove that fuzzy $\mathcal{ALC}$
under \L ukasiewicz enjoys the FMP since it is also known that
first-order formulae satisfiable under \L ukasiewicz semantics are
always satisfiable in a witnessed model (see~\cite{Hajek07}). On the
other hand, the FMP fails under G\"odel and Product semantics even
for ABoxes~\cite[Example~2]{Hajek05}.

The aim of this paper is to show that, contrary to what was
thought\footnote{And implicitly used in several published
algorithms (see Section~\ref{sec:discussion}).} up to now, the previous result cannot be extended to
arbitrary knowledge bases; that is, for arbitrary knowledge bases
satisfiability in a finite model may not coincide with
satisfiability in the class of witnessed models. In this paper we
show that this happens at least in the cases of \L ukasiewicz and
Product semantics. For these two cases we provide knowledge bases
that are satisfiable in a witnessed model but not in any finite
model. In particular, it follows that FMP is false in \L ukasiewicz
when we allow arbitrary knowledge bases.

The main reason why we can build knowledge bases that are
counterexamples to the FMP is the use of general concept
inclusions in the TBox. Indeed, in Section~\ref{sec:discussion} we
show that if the TBox is unfoldable then satisfiability in a finite
model coincides with satisfiability in a witnessed model.

The remainder of this document is as follows.
Section~\ref{sec:fuzzyALC} recalls the definition of fuzzy
$\mathcal{ALC}$s. Then, Section~\ref{sec:infiniteModels} shows that,
under \L ukasiewicz and Product semantics satisfiability of a
knowledge base in a witnessed model does not coincide with
satisfiability in a finite model. Finally,
Section~\ref{sec:discussion} discusses some consequences of this
fact.


\section{Fuzzy Description Logics $\mathcal{ALC}$}
\label{sec:fuzzyALC}

The crisp DL $\mathcal{ALC}$ is one particular logic. On the other
hand, its generalization to the fuzzy setting is not unique since it
depends on which fuzzy operators are chosen
(see~Table~\ref{tab:fuzzyOperators}). In this section, for every
family of operators it is defined a particular $\mathcal{ALC}$ fuzzy
logic based on the guidelines given
in~\cite{Straccia01,StracciaSHOIND}. It is worth pointing out that
all these fuzzy DLs share the same syntax, although their semantics
are different.

\paragraph{\textsc{Syntax}}

Let $\AC$, $\AR$ and $\AI$ be non-empty enumerable and pair-wise
disjoint sets of \emph{concept names} (denoted $A$), \emph{role
names} (denoted $R$) and \emph{individual names} (denoted $a, b$).
Concepts may be seen as unary predicates, while roles may be seen as
binary predicates.

$\mathcal{ALC}$ (complex) concepts can be built according to the
following syntax rule:

\[
\begin{array}{ccl}
C,D & \to & \topc \ | \ \bottomc  \ | \ A  \ | \ C \andc D \ | \
C \orc D \ | \ \notc C \ | \ \all R.C \ | \ \csome R.C \\
\end{array}
\]

There are no complex roles in $\mathcal{ALC}$, but only the atomic
ones. Thus, the syntax here considered is essentially the same one
than for crisp $\mathcal{ALC}$. Although it would be interesting to
increase the expressive power adding an implication $C \Rightarrow
D$ of concepts, we have adopted the convention to avoid it in order
to present a weaker logic where still the negative results given in
Section~\ref{sec:infiniteModels} hold.

An $\mathcal{ALC}$ \emph{fuzzy knowledge base} (fuzzy KB) $\KB =
\tuple{\A, \T}$ consists of a fuzzy ABox $\A$ and a fuzzy TBox $\T$,
where
\begin{itemize}
  \item a \emph{fuzzy ABox} $\A$ is a finite set of \emph{fuzzy
    concept assertion} axioms of the form $\fuzzyg{\cass{a}{C}}{\alpha}$
    and $\fuzzyneg{\cass{a}{C}}{\alpha}$, and \emph{fuzzy role
    assertion} axioms of the form\footnote{The reader may be surprised
    about the fact that axioms of the form
    $\fuzzyneg{\rass{a}{b}{R}}{\alpha}$ are not considered. The reason
    to not accept this kind of statements is that in the case of crisp
    $\mathcal{ALC}$ in the ABox it is not allowed to use statements of
    the form $\fuzzyneg{\rass{a}{b}{R}}{0}$ (remember that in
    $\mathcal{ALC}$ the negation of a role is not allowed). On the other
    hand, in crisp $\mathcal{ALC}$ it is indeed allowed to use
    $\fuzzyneg{\cass{a}{C}}{0}$ since it corresponds to
    $\fuzzyg{\cass{a}{\neg C}}{1}$ (remember that the negation of a
    concept is allowed). } $\fuzzyg{\rass{a}{b}{R}}{\alpha}$, where
    $a,b$ are individual names, $C$ is a concept, $R$ is a role and
    $\alpha \in [0,1]$ is a rational number.

  \item a \emph{fuzzy TBox} $\T$ is a finite set of \emph{fuzzy general concept
    inclusion} (GCI) axioms $\fuzzyg{C \impc D}{\alpha}$, where $C,D$ are
    concepts and $\alpha \in [0,1]$ is a rational number.
\end{itemize}

It is common to write $\tau$ as a shorthand for $\fuzzyg{\tau}{1}$,
$\langle \cass{a}{C} = \alpha \rangle$ as an abbreviation of the
pair of axioms $\fuzzyg{\cass{a}{C}}{\alpha}$ and
$\fuzzyneg{\cass{a}{C}}{\alpha}$, and the \emph{concept equivalence}
$C \equiv D$ as a shorthand of the two axioms $\fuzzyg{C \impc
D}{1}$ and $\fuzzyg{D \impc C}{1}$.

For the sake of concrete illustration, let us introduce a couple of
examples of fuzzy KBs. The second one will be used later as a
counterexample to the FMP.

\begin{example}
The pair
$\KB_1 = \tuple{\A_1, \T_1}$ with $\A_1 = \{
\fuzzyg{\cass{\mathtt{jim}}{\mathtt{YoungPerson}}}{0.2}$,
$\fuzzyg{\rass{\mathtt{jim}}{\mathtt{mary}}{\mathtt{likes}}}{0.8}
\}$ and $\T_1 = \{ \fuzzyg{Inn \sqsubseteq Hotel}{0.5} \}$ is a fuzzy KB.
\end{example}

\begin{example}
  \label{ex:KB}
$\K$ is the fuzzy KB with the following axioms
\begin{enumerate}
  \item[{\rm (1)}] $(a : A) = 0.5$
  \item[{\rm (2)}] $\top \sqsubseteq \exists R.\top$
  \item[{\rm (3)}] $(\forall R.A) \equiv (\exists R.A)$
  \item[{\rm (4)}] $A \equiv (\forall R.A) \sqcap (\forall R.A)$
\end{enumerate}

\end{example}


\paragraph{\textsc{Semantics}}

In fuzzy DLs, concepts and roles are interpreted, respectively, as
fuzzy subsets and fuzzy relations over an interpretation domain.
However, the axioms of a fuzzy KB (i.e., the elements of its ABox or
its TBox) are either satisfied (true) or unsatisfied (false) in an
interpretation. Hence, the axioms behave as in the crisp case and
they are not interpreted as a degree of truth in $[0,1]$.

Informally speaking, a fuzzy axiom $\fuzzyg{\ass}{\alpha}$ in a ABox constrains the
membership degree of $\ass$ to be at least $\alpha$. And the
intended semantics of $\fuzzyg{C \impc D}{\alpha}$ states that all
instances of concept $C$ are instances of concept $D$ to degree
$\alpha$, i.e. the subsumption degree (to be clarified later) between
$C$ and $D$ is at least $\alpha$. Next, we give precise definitions for
the semantics just explained.

A \emph{fuzzy interpretation (or model)} is a pair $\I =
(\highi{\Delta}, \highi{\cdot})$ consisting of a nonempty (crisp)
set $\highi {\Delta}$ (the \emph{domain}) and of a \emph{fuzzy
interpretation function\/} $\highi{\cdot}$ that assigns:
\begin{enumerate}

    \item to each atomic concept $A$ a function
      $\highi{A}\colon\highi{\Delta} \rightarrow \unit$,

    \item to each role $R$ a function $\highi{R}\colon\highi{\Delta}
      \times \highi{\Delta} \rightarrow \unit$,

    \item to each individual $a$ an element $\highi{a} \in \highi{\Delta}$.

\end{enumerate}

The fuzzy interpretation function is extended to 
complex concepts as specified in the Table~\ref{DLsem} (where $x,y
\in \highi{\Delta}$ are elements of the domain, and as usual we use
$\andf, \orf, \notf$ and $\impf$ to denote respectively the t-norm,
t-conorm, negation function and implication function of the
corresponding family of fuzzy operators chosen). Hence, for every
complex concept $C$ we get a function $C^{\I}: \highi{\Delta}
\longrightarrow [0,1]$.

\begin{table*}
\caption{Fuzzy DL semantics for ${\mathcal{ALC}}$.} \label{DLsem}
\[
\begin{array}{rcl}
\highi{\bottomc}(x) & = & 0 \\
\highi{\topc}(x) & = & 1 \\
\highi{(C \andc D)}(x) & = & \highi{C}(x) \andf \highi{D}(x)\\
\highi{(C \orc D)}(x) & = & \highi{C}(x) \orf \highi{D}(x) \\
\highi{(\notc C)}(x) & = & \notf \highi{C}(x)\\
\highi{(\all R.C)}(x) & = & \inf_{y \in \highi{\Delta}} (\highi{R}(x,y) \impf \highi{C}(y) ) \\
\highi{(\csome R.C)}(x) & = & \sup_{y \in \highi{\Delta}} (\highi{R}(x,y) \andf \highi{C}(y) ) \\
\end{array}
\]
\end{table*}

Note that $\fuzzyneg{\cass{a}{C}}{\alpha}$ is equivalent to
$\fuzzyg{\cass{a}{\neg C}}{1 - \alpha}$ under \L ukasiewicz
negation. Hence, in {\L}ukasiewicz it is not necessary to explicitly
consider fuzzy concept assertions of the form
$\fuzzyneg{\cass{a}{C}}{\alpha}$.

The \emph{(crisp) satisfiability of axioms in a fuzzy KB} is then defined by the
following conditions:
\begin{enumerate}
  \item $\I$ satisfies an axiom $\fuzzyg{\cass{a}{C}}{\alpha}$ in case
    that $C^{\I} (a^{\I}) \geq \alpha$,
  \item $\I$ satisfies an axiom $\fuzzyneg{\cass{a}{C}}{\alpha}$ in case
    that $C^{\I} (a^{\I}) \leq \alpha$,
  \item $\I$ satisfies an axiom $\fuzzyg{\rass{a}{b}{R}}{\alpha}$ in case
    that $R^{\I} (a^{\I}, b^{\I}) \geq \alpha$,
  \item $\I$ satisfies an axiom $\fuzzyg{C \impc D}{\alpha}$ in case
    that $\highi{(C \impc D)} \geq \alpha$ where $\highi{(C \impc D)}$ $=
    \inf_{x \in \highi{\Delta}} ( \highi{C}(x) \impf \highi{D}(x))$.

\end{enumerate}

It is interesting to point out that the satisfaction of a GCI of the
form $\fuzzyg{C \impc D}{1}$ is exactly the requirement that
$\forall x \in \Delta^{\I}, \highi{C}(x) \leq \highi{D}(x)$ (i.e.,
Zadeh's set inclusion); hence, in this particular case for the
satisfaction it only matters the partial order and not the exact
value of the implication $\Rightarrow$.

As it is expected we will say that a fuzzy interpretation $\I$
satisfies a KB $\KB$ in case that it satisfies all axioms in $\KB$.
And it is said that a fuzzy KB $\KB$ is \emph{satisfiable} iff there
exist a fuzzy interpretation $\I$ satisfying every axiom in $\KB$.


As it is said in the introduction, here we mainly focus on witnessed
models and compare them to finite models. This notion
(see~\cite{Hajek05}) corresponds to the restriction to the DL
language of the notion of witnessed model introduced, in the context
of the first-order language, by H\'ajek in \cite{Hajek07}. A fuzzy
interpretation $\I$ is said to be \emph{witnessed} iff it holds that
for every complex concepts $C, D$, every role $R$, and every $x \in
\highi{\Delta}$ there is some
\begin{enumerate}
\item $y \in \highi{\Delta}$ such that
$(\some R.C)^{\I}(x) = R^{\I}(x,y) \otimes C^{\I}(y)$.

\item $y \in \highi{\Delta}$ such that
$(\forall R.C)^{\I}(x) = R^{\I}(x,y) \Rightarrow C^{\I}(y)$.

\item $y \in \highi{\Delta}$ such that $(C \sqsubseteq D)^{\I} =
C^{\I}(y) \Rightarrow D^{\I}(y)$.
\end{enumerate}

The idea behind this definition is that in a witnessed
interpretation all arbitrary infima and suprema needed in order to
compute $C^{\I}$ and $(C \sqsubseteq D)^{\I}$ are indeed minima and
maxima. It is obvious that all finite fuzzy interpretations (this
means that $\highi{\Delta}$ is a finite set) are indeed witnessed,
but the opposite is not true. A fuzzy KB $\KB$ is said to be \emph{satisfiable
in a witnessed interpretation} iff there exist a witnessed fuzzy
interpretation $\I$ satisfying every axiom in $\KB$.


\section{Infinite Models in Fuzzy $\mathcal{ALC}$}
\label{sec:infiniteModels}

In this section we prove that there are fuzzy KBs that are not
finitely satisfiable while are satisfiable in a witnessed
interpretation. We will prove this statement for two cases: the \L
ukasiewicz and the Product cases, and in both cases we will use the
same fuzzy KB $\K$ defined in Example~\ref{ex:KB}.

First of all we focus in the \L ukasiewicz case.  It is worth pointing
out that under \L ukasiewicz semantics the first axiom in $\K$ can be
rewritten as \[ \fuzzyg{\cass{a}{( (A \orc A) \andc \notc (A \andc A)
)}}{1}, \] and so it is simply using as a bound the crisp value $1$.

\begin{proposition}
  \label{prop:Luk}

Let $\I$ be a witnessed model of $\K$ under \L ukasiewicz fuzzy logic. Then,
for every natural number $n$ there are individuals $b_1, b_2, b_3,
\dots, b_n$ such that $0.5 = A^{\I}(b_1) < A^{\I}(b_2) < A^{\I}(b_3)
< \dots < A^{\I}(b_n) < 1$.
\end{proposition}

\begin{pf}
For the case $n = 1$ this is trivial using axiom (1) in the fuzzy
KB.

Now let us assume that there are individuals $b_1, b_2, b_3, \dots,
b_n$ such that $0.5 = A^{\mathcal{I}}(b_1) < A^{\mathcal{I}}(b_2) <
A^{\mathcal{I}}(b_3) < \dots < A^{\mathcal{I}}(b_n) < 1$. We want to
check that there is an individual $b_{n+1}$ such that
$A^{\mathcal{I}}(b_n) < A^{\mathcal{I}}(b_{n+1}) < 1$.

Using axiom (2) we know that $(\exists R.\top)^{\mathcal{I}}(b_n) =
1$. Since the model is witnessed, there is a
$1$-successor of $b_n$ that we call $b$, i.e.,
$R^{\mathcal{I}}(b_n, b) = 1$. It is obvious that $(\forall
R.A)^{\mathcal{I}}(b_n) \leq 1 \Rightarrow A^{\mathcal{I}}(b) =
A^{\mathcal{I}}(b)$ and that $A^{\mathcal{I}}(b) = 1 \otimes
A^{\mathcal{I}}(b) \leq (\exists R.A)^{\mathcal{I}}(b_n)$. Using
axiom (3) we get that $(\forall R.A)^{\mathcal{I}}(b_n) =
A^{\mathcal{I}}(b) =(\exists R.A)^{\mathcal{I}}(b_n)$. Therefore,
axiom (4) is saying that $A^{\mathcal{I}}(b_n) = (\forall
R.A)^{\mathcal{I}}(b_n) \otimes (\forall R.A)^{\mathcal{I}}(b_n) =
A^{\mathcal{I}}(b) \otimes A^{\mathcal{I}}(b)$. Using \L ukasiewicz
t-norm, it follows that $A^{\mathcal{I}}(b_n) = A^{\mathcal{I}}(b) +
A^{\mathcal{I}}(b) - 1$, and hence $A^{\mathcal{I}}(b) =
(A^{\mathcal{I}}(b_n) + 1) / 2$. Thus, it is clear that $A(b_n) <
A(b) < 1$. Hence, we can define $b_{n+1}$ as $b$ to finish
the proof. \qed
\end{pf}

\begin{corollary}
There is no finite model for $\K$ under \L ukasiewicz
fuzzy logic.
\end{corollary}

\begin{theorem}
  \label{thm:Luk}
  $\K$ is, under \L ukasiewicz fuzzy logic, satisfiable by a witnessed
  model but not by a finite model.
\end{theorem}

\begin{pf}
  By the previous corollary it is enough to give a
  witnessed model of $\K$. One witnessed model of this fuzzy KB is the
  model $\I$ defined by
\begin{itemize}
    \item $\Delta^{\I} = \{1, 2, 3, \dots \} \cup \{
      \infty \}$,
    \item $R^{\I}$ is the crisp relation 
      $\{({i},{i+1}) : i = 1, 2, 3, \dots \} \cup \{ (\infty,\infty) \}$,
    \item $A^{\I}(\infty) = 1$ and $A^{\I}(i) = (2^i
      - 1) / 2^i$ for every $i=1,2,3,\dots$
    \item $a^{\I} = 1$.
\end{itemize}
  The fact that this model $\I$ satisfies $\K$ can be easily checked by
  the reader, and the same for the first two conditions in the
  definition of witnessed model. On the other hand, the last condition
  in the definition of witnessed model is subtler, and we give here a
  proof. First of all we notice the following claim.

\begin{claim}
  For every (complex) concept $C$ and every $\varepsilon > 0$, it holds one (and
only one) of the following two conditions:
\begin{description}
  \item[$\boldsymbol{Cond_1(C, \varepsilon)}$:] $C^{\I}(\infty) = 0$,
    and there is some $n \in \mathbb{N}$ such that $\inf \{C^{\I}(i) :i
    \in \mathbb{N}, i \geq n \} = 0$ and $\sup \{C^{\I}(i) :i \in
    \mathbb{N}, i \geq n \} \leq \varepsilon$.
  \item[$\boldsymbol{Cond_2(C, \varepsilon)}$:] $C^{\I}(\infty) = 1$,
    and there is some $n \in \mathbb{N}$ such that $\sup \{C^{\I}(i) :i
    \in \mathbb{N}, i \geq n \} = 1$ and $\inf \{C^{\I}(i) :i \in
    \mathbb{N}, i \geq n \} \geq 1 - \varepsilon$.
\end{description}
\end{claim}

\begin{pfclaim}
  First of all, we notice that a trivial induction proves that
  $C^{\I} (\infty) \in \{ 0, 1 \}$ for every concept $C$.
  Next we prove the claim by induction on the length of the concept
  $C$. The cases of $\top$ and an atomic concept $A$ are trivial. Due to the
  interdefinability between connectives in the {\L}ukasiewicz case, it
  is enough to study the connectives $\neg$, $\sqcap$ and $\forall R.C$.
  \begin{itemize}
    \item The case of a concept $\neg C$ follows from the continuity of
      the {\L}ukasiewicz negation; we remind the reader that continuity tells us
      that the function commutes with infima and suprema.

    \item Let us consider the case of a concept $C \sqcap D$. If both
      $C$ and $D$ satisfy the second condition, then it is enough to see
      that $Cond_2(C, \varepsilon/2)$ together with $Cond_2(D,
      \varepsilon/2)$ implies $Cond_2(C \sqcap D, \varepsilon)$; and
      this fact is a consequence of
      $\sup \{ (C \sqcap D)^{\I}(i) :i \in
      \mathbb{N}, i \geq n \} = \sup \{C^{\I}(i) :i \in \mathbb{N}, i
      \geq n \} \otimes \sup \{D^{\I}(i) :i \in \mathbb{N}, i \geq n \} =
      1 \otimes 1= 1$ (by the continuity
      of $\otimes$) and $(1 - \frac{\varepsilon}{2}) \otimes (1 -
      \frac{\varepsilon}{2}) \geq 1 - \varepsilon$.
      Let us consider now the case that at least one of the
      two concepts satisfies the first condition. Without loss of
      generality, we can assume that $Cond_1(C, \varepsilon)$ holds.
      From this assumption it follows $Cond_1(C \sqcap D, \varepsilon)$
      because $(C \sqcap D)^{\I}(i) \leq C^{\I} (i)$ for every $i$.

    \item The case of a concept $\forall R.C$ follows from the fact
      that for every $i \in \mathbb{N}$, $(\forall R. C)^{\I} (i) =
      C^{\I} (i+1)$.
  \end{itemize}
  This finishes the proof of the claim.
\hfill{\footnotesize\textsc{q.e.d.}}
\end{pfclaim}

Let us now prove the third requirement in the definition of a witnessed
interpretation. Hence, we consider two concepts $C$ and $D$. Using the
previous claim for the concept $\neg C \sqcup D$ we are in one of the
following two cases.
\begin{itemize}
  \item Case $(\neg C \sqcup D)^{\I} (\infty) = 0$. In this case it is
    trivial that $(C \sqsubseteq D)^{\I} = 0 = C^{\I}(\infty)
    \Rightarrow D^{\I} (\infty)$.

  \item Case $(\neg C \sqcup D)^{\I} (\infty) = 1$ and for every
    $\varepsilon > 0$ there is some $n
    \in \mathbb{N}$ such that $\sup \{(\neg C \sqcup D)^{\I}(i) :i \in
    \mathbb{N}, i \geq n \} = 1$ and $\inf \{(\neg C \sqcup D)^{\I}(i) :i \in
    \mathbb{N}, i \geq n \} \geq 1 - \varepsilon$. Without loss of
    generality we can assume that
    there is some $m \in \mathbb{N}$ such that $(\neg C
    \sqcup D)^{\I} (m) < 1$, because if not, then $(C \impc D)^{\I} = 1$
    and hence $(C \impc D)^{\I} = C^{\I}(i) \impf D^{\I}(i)$
    for every $i \in I$. Let us consider $a:=(\neg C
    \sqcup D)^{\I} (m)$  and $\varepsilon:= 1 - a$.
    Thus, using the claim we know that there is some $n \in \mathbb{N}$ such that $\sup
    \{(\neg C \sqcup D)^{\I}(i) :i \in \mathbb{N}, i \geq n \} = 1$ and
    $\inf \{(\neg C \sqcup D)^{\I}(i) :i \in \mathbb{N}, i \geq n \} \geq a$. Hence,
    $$
    (C \sqsubseteq D)^{\I} = \min ( \{ (\neg C \sqcup D)^{\I}(m) \} \cup
    \{ (\neg C \sqcup D)^{\I}(i): i \leq n \} ).
    $$
    As $(C \sqsubseteq D)^{\I}$ is the minimum of a finite set the proof is
    finished.
  \qed
\end{itemize}
\end{pf}


Next we consider the case of Product logic. The following
proposition can be proved by essentially the same argument as for
Proposition~\ref{prop:Luk}, so we skip the proof.

\begin{proposition}
  \label{prop:product}
Let us a consider a witnessed model $\I$ of $\K$ under Product fuzzy logic.
Then, for every natural number $n$ there are individuals $b_1, b_2,
b_3, \dots, b_n$ such that $0.5 = A^{\I}(b_1) < A^{\I}(b_2) <
A^{\I}(b_3) < \dots < A^{\I}(b_n) < 1$.
\end{proposition}

\begin{corollary}
There is no finite model for $\K$ under Product fuzzy logic.
\end{corollary}

\begin{theorem}
  \label{thm:product}
  $\K$ is, under Product fuzzy logic, satisfiable by a witnessed
  model but not by a finite model.
\end{theorem}

\begin{pf}
  It is enough to give a witnessed model of $\K$. One witnessed model of this fuzzy KB is the
  model $\I$ defined by
\begin{itemize}
    \item $\Delta^{\I} = \{ 1, 2, 3, \dots \}$,

    \item $R^{\I}$ is the crisp relation $\{({i},{i+1}) : i = 1, 2, 3, \dots \}$,


    \item $A^{\I}(i) = \sqrt[2^{i-1}]{\big(\frac{1}{2}\big)}$ for
      every $i = 1,2, 3, \dots$,

    \item $a^{\I} = 1$.
\end{itemize}
The fact that this model $\I$ satisfies $\K$ can be easily checked
by the reader, and the same for the first two conditions in the
definition of witnessed model. The rest of this proof is devoted to
prove the third condition.

\begin{claim}
  For all (complex) concepts $C$, one (and
only one) of the following conditions holds:
\begin{itemize}
  \item $0 = C^{\I}(i)$ for every $i \in \mathbb{N}$,
  \item $0 < C^{\I} (1) \leq C^{\I} (2) \leq C^{\I} (3) \leq \ldots$ and
    $\sup \{C^{\I}(i) :i \in \mathbb{N} \} = 1$.
\end{itemize}
\end{claim}

\begin{pfclaim}
This can be straightforwardly proved by induction on the length of
the concept $C$ using how the fuzzy operators involved in the
product case are defined\footnote{Here, as opposed to the \L
ukasiewicz case, the fact that the implication operator
$\Rightarrow$ is not involved in the construction of complex
$\mathcal{ALC}$ concepts is crucial.}.
\begin{itemize}
  \item The case of a concept $\neg C$ is a trivial consequence of the
    fact that Product negation is the one given in
    Table~\ref{tab:fuzzyOperators} (usually called strict negation).
  \item The cases of concepts built with $\sqcap$ and $\sqcup$ follow
    from the monotonicity of the functions $\otimes$ and $\oplus$.
  \item The cases of concepts built with $\forall R.$ and $\exists R.$ follow
    from the fact that for every $i \in \mathbb{N}$, $(\forall R.C)^{\I} (i) = C^{\I} (i+1 ) =
    (\exists R.C)^{\I} (i)$.
\end{itemize}
This finishes the proof of the claim.
\hfill{\footnotesize\textsc{q.e.d.}}
\end{pfclaim}

Now it is time to use this previous claim in order to check that
for every pair of concepts $C$ and $D$ there is some $i \in
\mathbb{N}$ such that $(C \impc D)^{\I} =C^{\I}(i) \Rightarrow
D^{\I}(i)$. By the previous property we can distinguish three
different cases, which cover all possibilities.
\begin{itemize}
  \item Firstly, we consider the case where $0 = C^{\I}(i)$ for
    every $i \in \mathbb{N}$. Then, it is trivial that $(C \impc D)^{\I}
    = \inf \{ 0 \impf D^{\I}(i) : i \in \mathbb{N} \} = 1 = C^{\I}(i) \impf D^{\I}(i)$
    for every $i \in \mathbb{N}$.

  \item Secondly, we consider the case where  $0 \neq C^{\I}(i)$
and $0 = D^{\I}(i)$ for every $i \in \mathbb{N}$. Then, it is
trivial that $(C \impc D)^{\I} = \inf \{ C^{\I}(i) \impf 0 : i \in
\mathbb{N} \} = 0 = C^{\I}(i) \impf D^{\I}(i)$ for every $i \in
\mathbb{N}$.

  \item Lastly, we consider the case where  $0 < C^{\I} (1) \leq C^{\I}(2)
  \leq C^{\I}(3) \leq \ldots$, $\sup \{C^{\I}(i) :i \in \mathbb{N}
    \} = 1$, $0 < D^{\I}(1) \leq D^{\I}(2) \leq D^{\I}(3) \leq \ldots$ and
    $\sup \{D^{\I}(i) :i \in \mathbb{N} \} = 1$. In
    order to finish this proof it suffices to consider the case that
    $(C \impc D)^{\I} \neq 1$, because if $(C \impc D)^{\I} = 1$ then it
    is clear that $(C \impc D)^{\I} = C^{\I}(i) \impf D^{\I}(i)$
    for every $i \in I$. Hence, let us assume that $(C \impc D)^{\I}
    \neq 1$. Thus, there is some $i_0 \in \mathbb{N}$ such that $(C^{\I}
    (i_0) \impf D^{\I} (i_0)) = \frac{D^{\I}(i_0)}{C^{\I}(i_0)}
    < 1$.
    On the other hand, by continuity we know that
    \[
    \lim_{i \to \infty} \min \left\{ \frac{C^{\I}(i)}{D^{\I}(i)},
    \frac{D^{\I}(i)}{C^{\I}(i)} \right\}
    \: = \:
    \min \left\{ \frac{\lim_{i \to \infty} C^{\I}(i)}{\lim_{i \to \infty} D^{\I}(i)},
    \frac{\lim_{i \to \infty} D^{\I}(i)}{\lim_{i \to \infty} C^{\I}(i)}
    \right\}
    \: = \:
    \]
    \[
    \: = \:
    \min \left\{ \frac{\sup_{i \in \mathbb{N}} C^{\I}(i)}{\sup_{i \in \mathbb{N}}
    D^{\I}(i)}, \frac{\sup_{i \in \mathbb{N}} D^{\I}(i)}{\sup_{i \in \mathbb{N}}
    C^{\I}(i)} \right\}
    \: = \:
    1
    \]
    Therefore, as a consequence of the fact that for every $i \in \mathbb{N}$,
    \[
    \min \left\{ \frac{C^{\I}(i)}{D^{\I}(i)},
    \frac{D^{\I}(i)}{C^{\I}(i)} \right\}
    \: \leq \: C^{\I}(i) \impf D^{\I}(i) \: \leq \: 1,
    \]
    it follows that $\lim_{i \to \infty} (C^{\I}(i)
    \impf D^{\I}(i)) = 1$.
    Thus, we know that the set $I =
    \{ i \in \mathbb{N}: C^{\I}(i) \impf D^{\I}(i) \: \leq \:
    C^{\I}(i_0) \impf D^{\I} (i_0)\}$ is finite. Since $I$ is
    finite it is obvious that there is some $i \in I$ such that $(C
    \impc D)^{\I} = C^{\I}(i) \impf D^{\I}(i)$.
  \qed
\end{itemize}
\end{pf}


\section{Discussion} \label{sec:discussion}

The aim of this section is to discuss some interesting remarks of
the failure of the FMP exposed in Theorems~\ref{thm:Luk}
and~\ref{thm:product}.

The first remark is that the failure of the FMP is essentially a
consequence of the fact that the TBox is \emph{cyclic}. As we will
see, in the case of {\L}ukasiewicz this situation does not happen if
we consider only unfoldable KBs.

A TBox $\T$ is \emph{acyclic} iff it verifies the following three
constraints:
\begin{enumerate}
\item Every axiom in $\T$ is either of the form
$\fuzzyg{A \impc C}{\alpha}$ or $A \equiv C$, where $A$ is an atomic
concept.\footnote{In particular this forbids concept $\top$ on the
left hand side of axioms.}

\item There is no concept $A$ such that it appears more than once on the
left hand side of some axiom in $\T$.

\item No cyclic definitions are present in $\T$.
We will say that $A$ \emph{directly uses} primitive concept $B$ in
$\mathcal{T}$, if there is some axiom $\tau \in \mathcal{T}$ such
that $A$ is on the left hand side of $\tau$ and $B$ occurs in the
right hand side of $\tau$. Let \emph{uses} be the transitive closure
of the relation \emph{directly uses} in $\mathcal{T}$. $\mathcal{T}$
is cyclic iff there is $A$ such that $A$ uses $A$ in $\mathcal{T}$.

\end{enumerate}
Analogously, a fuzzy KB is said to be
\emph{acyclic} when its TBox is acyclic.

\medskip

In the crisp case, acyclic TBoxes can be eliminated through an
\emph{expansion process}~\cite{Nebel90} which only takes a finite
number of steps. The fact that this process can create an
exponential growth of the KB has motivated the introduction of the
\emph{lazy expansion} optimization technique~\cite{LazyExpansion},
very useful in practice.

In the fuzzy case, acyclic TBoxes cannot be completely removed in
general due to the presence of degrees $\alpha$, and we need to
require some additional property in order to do so. We say that a
fuzzy TBox is \emph{unfoldable} if it is an acyclic TBox which only
contains axioms of the form $\fuzzyg{A \impc C}{1}$, and $A \equiv
C$ (that is, an acyclic TBox where every fuzzy concept inclusion
axiom is of the form $\fuzzyg{A \impc C}{\alpha}$ with $\alpha = 1$)
(cf.~\cite[Section~3.3]{Straccia01}). Analogously, a fuzzy KB is
said to be \emph{unfoldable} when its TBox is unfoldable.

In case we allow to use truth constants $\alpha$ as concept
constructors, then it is trivially true for all left-continuous
t-norms that $\fuzzyg{C \impc D}{\alpha}$ is equisatisfiable with
$\fuzzyg{C \sqcap \alpha \impc D}{1}$.  Next, we exploit a similar
idea to show that, in the particular case of {\L}ukasiewicz,
unfoldable fuzzy TBoxes can be converted into acyclic ones, even
without having truth constants $\alpha$ as concept constructors. The
idea is that all axioms of the form $\fuzzyg{A \impc C}{\alpha}$ can
be converted into the form $\fuzzyg{A \impc C'}{1}$. For example,
$\fuzzyg{A \impc C}{\frac{2}{3}}$ can be converted into the form
\[
\fuzzyg{A \impc C \sqcup ((A' \sqcap A') \sqcap \neg (A' \sqcap A' \sqcap
A'))}{1} \ ,
\]
\nd where $A'$ is a new atomic concept; the reason why this is so is because the concept $(A'
\sqcap A') \sqcap \neg (A' \sqcap A' \sqcap A')$ always takes values onto the
interval $[0, \frac{1}{3}]$. In the following proof we develop this idea.

\begin{lemma}[{\L}ukasiewicz Case]
\label{lem:acyclicLuk}
  There is an algorithm that converts every acyclic fuzzy KB $\KB =
  \tuple{\A, \T}$ into an unfoldable fuzzy KB $\KB' =
  \tuple{\A, \T'}$ in such a way that
  \begin{center}
    $\KB$ is satisfiable \qquad iff \qquad $\KB'$ is satisfiable.
  \end{center}
  The algorithm also preserves satisfiability in a witnessed model, and
  the same for satisfiability in a finite model.
\end{lemma}

\begin{pf}
  Let us consider an axiom of the form $\fuzzyg{A \impc C}{\alpha}$. By
  the constructive version of McNaughton's Theorem (see~\cite{Agu98}) it
  is obvious that there is an algorithm that for every rational
  $\alpha$ generates a propositional concept $\tau_{\alpha}(A')$ using
  only one atomic concept $A'$ and such that
  \begin{description}
    \item[P1:] $\tau_{\alpha}(A')$ only takes values in $[0, 1 - \alpha]$,
    \item[P2:] $\tau_{\alpha}(A')$ takes value $1- \alpha$ when $A'$ takes value
      $\alpha$.
  \end{description}
  To finish the proof it is enough to see that for every KB $\KB$, it
  holds that
  \begin{itemize}
    \item $\KB \cup \{ \fuzzyg{A \impc C}{\alpha} \}$ is satisfiable,
      iff
    \item $\KB \cup \{ \fuzzyg{A \impc C \sqcup \tau_{\alpha}(A')}{1}
      \}$ is satisfiable (where $A'$ is a new atomic concept).
  \end{itemize}
  The upwards implication is a consequence of P1. On the other hand, the
  downwards implication is a consequence of P2 because we can extend any
  interpretation $\I$ to encompass the new atomic concept $A'$ defining
  for every individual $x$, $A'^{\I} (x) = \alpha$. It is also obvious
  that this way of extending the interpretation preserves
  the witnessed property (because $A'$ takes the same value in all
  individuals) and the finiteness.
  \qed
\end{pf}

Next we prove, using ideas from \cite{Straccia01}, that fuzzy
unfoldable TBoxes can be eliminated through a finite expansion
process when the interpretation of $\sqcap$ is a continuous t-norm.
Our proof of this result does not seem to work for satisfiability in
a witnessed model; but we know how to deal with this case as long as
the minimum constructor $\min \{ C,D\}$ is definable in our
framework.  We remind the reader that in the case of continuous
t-norms the constructor $\min \{ C,D\}$ is definable, using an
R-implication constructor $\to$, by $C \sqcap (C \to D)$; hence, the
Zadeh, {\L}ukasiewicz (note that in \L ukasiewicz logic, $C
\rightarrow D$ coincides with $\notc C \orc D$) and G\"odel families
given in Table~\ref{tab:fuzzyOperators} satisfy the assumption in
the second item of the following result.

\begin{lemma}[Assuming $\otimes$ is continuous]
  \label{lem:equisatisfiability}
  Let $\KB$ be a fuzzy KB and let $C,D$ be two concepts.
  \begin{enumerate}
    \item The following statements are equivalent:
      \begin{itemize}
    \item $\mathcal{K} \cup \{ \fuzzyg{C \impc D}{1} \}$ is satisfiable,
    \item $\mathcal{K} \cup \{ C \equiv A \sqcap D \}$, where
      $A$ is a new atomic concept, is satisfiable.
      \end{itemize}
      This equivalence also holds when we consider satisfiability in a finite model.

    \item (Assuming the constructor $\min$ is definable) The following
      statements are equivalent:
      \begin{itemize}
    \item $\mathcal{K} \cup \{ \fuzzyg{C \impc D}{1} \}$ is satisfiable,
    \item $\mathcal{K} \cup \{ C \equiv \min \{ A , D \} \}$, where
      $A$ is a new atomic concept, is satisfiable.
      \end{itemize}
      This equivalence also holds when we consider satisfiability in a
      finite model, and the same for satisfiability in a witnessed model.
  \end{enumerate}
\end{lemma}

\begin{pf}
  Let us consider the first item.
  One direction is a consequence of the fact that t-norms
  satisfy $x \otimes y \leq y$. For the
  other, let us point out that continuous t-norms are divisible (see
  \cite[Lemma~2.1.10]{Hajek05}) in
  the sense that for every $x,y \in
  [0,1]$ there is an element $a$ such that $a \otimes x = \min \{ x,y\}$; this
  element $a$ can be defined as $\sup \{ z \in [0,1]: x \otimes z \leq y
  \}$ and is commonly denoted $x \Rightarrow y$. Hence, we can extend
  any interpretation $\I$ to encompass the new atomic concept $A$
  defining
  $A^{\I} (x) := D^{\I}(x) \Rightarrow C^{\I}(x)$.

  One direction of the second item follows from the fact that $\min\{x ,
  y \} \leq y$. The other is proved extending
  any interpretation $\I$ to encompass the new atomic concept $A$
  defining $A^{\I} (x) := C^{\I}(x)$.
  \qed
\end{pf}

\begin{lemma}[Assuming $\otimes$ is continuous] \textrm{}
  \label{lem:eliminateTBox}
  \begin{enumerate}
    \item There is an algorithm that converts every unfoldable fuzzy KB $\KB =
      \tuple{\A, \T}$ into a fuzzy ABox
      $\A^{\star}$ in such a way that
      \begin{center}
    $\KB$ is satisfiable \qquad iff \qquad $\A^{\star}$ is satisfiable.
      \end{center}
      The algorithm also preserves satisfiability in a finite model.

    \item (Assuming the constructor $\min$ is definable) There is
      an algorithm like in the previous item which additionally also
      preserves satisfiability in a witnessed model.
  \end{enumerate}
 \end{lemma}

\begin{pf}
  We give the proof of the first item. The second one is proved
  analogously but using the second item in
  Lemma~\ref{lem:equisatisfiability}.

  Let us assume that $\KB = \tuple{\A, \T}$ is an unfoldable fuzzy KB.
  We know that all axioms in $\T$ are either of the form $\fuzzyg{A
  \impc C}{1}$ or $A \equiv C$, where $A$ is an atomic concept.
  The algorithm consists on two steps. In the first step, we replace all
  axioms of the form $\fuzzyg{A \impc C}{1}$ with $A
  \equiv A' \sqcap C$ (with $A'$ a new atomic concept). Once this is
  done we know that all axioms in the new TBox $\KB'$ are of the form $A \equiv C$.
  In the second step of the algorithm, we consider the ABox $\A$ and
  replace, for every axiom $A \equiv C$, all occurrences of $A$ with
  $C$. We will refer by $\A^{\star}$ to the output of this process.

  Now let us check that $\KB$ is satisfiable iff $\A^{\star}$ is
  satisfiable. The first item in Lemma~\ref{lem:equisatisfiability}
  takes care of the first step of the algorithm, that is, $\KB$ is
  satisfiable iff $\KB'$
  is satisfiable. Next we prove that $\KB'$ is satisfiable iff
  $\A^{\star}$ is satisfiable. The rightwards direction is trivial. For
  the other, it is enough to notice that since the axioms in $\KB'$ are
  of the form $A \equiv C$ with $A$ an atomic concept we can redefine,
  for every individual $a$ appearing in the ABox $\A^{\star}$,
  $A^{\I} (\highi{a})$ as the value $C^{\I} (\highi{a})$.
\qed
\end{pf}

Next theorem is formulated only for \L ukasiewicz in order to use
the decidability result in~\cite[Corollary~1]{Hajek06b} for our
framework. We include t-conorm as a concept constructor as opposed
to~\cite{Hajek06b}, but \L ukasiewicz  t-conorm is clearly definable
from \L ukasiewicz t-norm and negation. Extensions of
~\cite[Corollary~1]{Hajek06b} adding a t-norm constructor would be
needed in order to deal with all t-norm logics considered in this
paper, and not only \L ukasiewicz.

\begin{theorem}[{\L}ukasiewicz Case]
\label{theor:acyclicLuk} \textrm{}
\begin{enumerate}
  \item For every fuzzy KB $\KB = \tuple{\A, \T}$ such that $\T$ is
    acyclic, the satisfiability in a finite model under the fuzzy
    $\mathcal{ALC}$ coincides with the satisfiability.

  \item There is an algorithm for checking the satisfiability of acyclic KBs.
\end{enumerate}
\end{theorem}

\begin{pf}
  By Lemma~\ref{lem:acyclicLuk} and Lemma~\ref{lem:eliminateTBox} it is possible to
  remove the TBox. Then, we use the known results~\cite[Theorem~8 and Corollary~1]{Hajek06b} that in
  fuzzy $\mathcal{ALC}$ ABoxes, satisfiability in a witnessed model and satisfiability in a
  finite model coincide and are decidable problems. Note
  that~\cite{Hajek06b} does not consider a t-conorm in the language, but it can
  defined from \L ukasiewicz negation and t-norm. The proof
  finishes using the result in~\cite{Hajek07} which
  tells us that for {\L}ukasiewicz, satisfiability coincides with
  satisfiability in a witnessed model. \qed
\end{pf}

On the other hand, the second remark concerns the fact there are
several reasoning algorithms in the literature for \L
ukasiewicz~\cite{BobilloIPMU2008,BobilloEusflat,BobilloMathware},
Product~\cite{BobilloProduct}, or any left continuous t-norm fuzzy
DLs~\cite{BobilloFSS,StoilosDL2009}) that claim to support GCIs.
These algorithms restrict themselves to witnessed models.
Unfortunately, these papers are implicitly assuming that the logic
satisfies FMP. By Theorems~\ref{thm:Luk} and~\ref{thm:product} we
have shown that this assumption cannot be done, so these algorithms
do not work correctly for arbitrary fuzzy KBs (even if the semantics
of GCIs is defined using Zadeh's set inclusion, as it happens in
~\cite{StoilosDL2009}). For instance, these algorithms are not able
to provide a correct model for $\K$. Once these algorithms generate
a tableau, it is unknown how to build an infinite model starting
from it.

By the previous results, the cited algorithms are correct if we add
some additional restrictions:

\begin{itemize}

    \item\cite{BobilloIPMU2008,BobilloEusflat,BobilloMathware} consider
    the fuzzy DL $\mathcal{ALC}$ under \L ukasiewcz fuzzy logic.
    According to Theorem~\ref{theor:acyclicLuk}, the algorithms are correct
    in case we only consider acyclic KBs.

    \item\cite{BobilloProduct} consider
    the fuzzy DL $\mathcal{ALC}$ under Product fuzzy logic.
    By Lemma~\ref{lem:eliminateTBox}, the algorithm is correct
    if we only consider unfoldable KBs.

    \item\cite{BobilloFSS} provides reasoning algorithms for
the fuzzy DLs $\mathcal{ALC}$ defined by families of fuzzy operators
corresponding to a left-continuous t-norm extended with an
involutive negation\footnote{Here the fuzzy logic corresponding to a
left continuous t-norm $\otimes$ is understood as based on the
connectives given by the t-norm $\otimes$, its residuum, \L
ukasiewicz negation and the dual t-conorm of $\otimes$. This is
quite different to the logic of left-continuous t-norms as used
in~\cite{Hajek98,Hajek05,Hajek06b}, and closer (except for the use
of truth constants as concept constructors) to the framework
in~\cite{EstevaIJAR}.}. This work restricts itself to acyclic KBs,
so for instance $\K$ cannot be represented in the logic. According
to Theorem~\ref{theor:acyclicLuk}, the algorithm is correct for \L
ukasiewicz logic. By Lemma~\ref{lem:eliminateTBox}, the algorithm is
also correct for unfoldable KBs. The correctness of the algorithm
for acyclic KBs any fuzzy logic different from \L ukasiewicz remains
a conjecture.

    \item\cite{StoilosDL2009} provides a reasoning algorithm for
the fuzzy DLs $\mathcal{SI}$ defined by families of fuzzy operators
corresponding to a left-continuous t-norm. Again, the algorithm is
correct if we restrict to unfoldable KBs
(Lemma~\ref{lem:eliminateTBox}), or to acyclic KBs and \L ukasiewicz
fuzzy logic (Theorem~\ref{theor:acyclicLuk}).

\end{itemize}

It is a matter of future research to determine the decidability of
fuzzy $\mathcal{ALC}$ and sublanguages with an unrestricted TBox
under \L ukasiewicz and Product fuzzy logics.


\section*{Acknowledgement}

We would like to thank to the anonymous referees for their very
valuable comments on an earlier version of this paper.  Special
acknowledgements are due to Marco Cerami, Francesc Esteva and
\`{A}ngel Garc\'ia-Cerda\~na for their comments on a preliminary
version.

\section*{Funding}

Fernando Bobillo has been partially funded by the Spanish Ministry
of Science and Technology (project TIN2009-14538-C02-01). F\'elix
Bou thanks Eurocores (LOMOREVI Eurocores Project
FP006/FFI2008-03126-E/FILO), Spanish Ministry of Education and
Science (project MULOG2 TIN2007-68005-C04-01), and Catalan
Government (2009SGR-1433).


\bibliographystyle{plain}

\end{document}